\documentclass[10pt,twocolumn,letterpaper]{article}

\usepackage[pagenumbers]{cvpr} 

\usepackage{multirow}
\usepackage{epsfig}
\usepackage{graphicx}
\usepackage{amsmath}
\usepackage{amssymb}
\usepackage{multirow}
\usepackage{graphics}
\usepackage{threeparttable}
\usepackage{color}
\usepackage[normalem]{ulem}
\usepackage{float}
\usepackage{amsfonts}
\usepackage{bm}
\usepackage{bbm}
\usepackage{enumitem}
\usepackage{natbib}
\usepackage{subcaption}
\usepackage{array}
\usepackage{colortbl}
\usepackage{pifont}
\usepackage{booktabs}
\usepackage{mathtools}
\usepackage{siunitx}
\usepackage{caption} 
\usepackage{xcolor}
\usepackage{subcaption} 
\usepackage[nopar]{lipsum}
\usepackage{listings}
\usepackage[export]{adjustbox}
\usepackage{makecell}

\usepackage{mathtools}
\usepackage{siunitx}
\usepackage[nopar]{lipsum}
\usepackage{listings}
\usepackage{xargs}                      
\usepackage{xcolor}
\usepackage{wrapfig} 
\definecolor{myy}{RGB}{126,95,0}
\definecolor{mygray}{gray}{.9}
\definecolor{Gray}{gray}{0.9}
\definecolor{bblue}{RGB}{30,80,120}
\definecolor{mygray1}{gray}{.7}
\definecolor{ggray}{RGB}{127,127,127}
\definecolor{defaultcolor}{gray}{.9}
\definecolor{dark-gray}{gray}{0.20}

\definecolor{mygreen}{HTML}{39b54a}

\newcolumntype{x}[1]{>{\centering\arraybackslash}p{#1pt}}
\newcolumntype{y}[1]{>{\raggedright\arraybackslash}p{#1pt}}
\newcolumntype{z}[1]{>{\raggedleft\arraybackslash}p{#1pt}}

\newlength\savewidth

\usepackage[dvipsnames,table]{xcolor}

\usepackage[dvipsnames,table]{xcolor}
\usepackage{arydshln}
\usepackage{times}
\usepackage{epsfig}
\usepackage{graphicx}
\usepackage{amsmath}
\usepackage{amssymb}
\usepackage{multirow}
\usepackage{algorithmic}
\usepackage{caption}
\usepackage{multicol}
\usepackage{array}
\usepackage{booktabs}
\usepackage{threeparttable}
\usepackage{color}
\usepackage{tcolorbox}
\definecolor{iccvblue}{rgb}{0.21,0.49,0.74}
\usepackage[pagebackref,breaklinks,colorlinks,allcolors=iccvblue]{hyperref}
\usepackage[capitalize]{cleveref}
\usepackage{pifont}
\usepackage{bbding}
\crefname{section}{Sec.}{Secs.}
\Crefname{section}{Section}{Sections}
\crefname{table}{Table}{Tables}
\crefname{table}{Tab.}{Tabs.}

\newcolumntype{P}[1]{>{\centering\arraybackslash}p{#1}}

\definecolor{mygray}{gray}{.9}
\definecolor{ggray}{RGB}{127,127,127}
\definecolor{reda}{RGB}{192,0,0}
\definecolor{redb}{RGB}{217,148,143}
\definecolor{myyellow}{RGB}{190,144,0}
\definecolor{mygreen}{RGB}{80,100,40}
\definecolor{myblue}{RGB}{30,90,100}
\definecolor{dark-gray}{gray}{0.20}
\definecolor{middle-gray}{gray}{0.85}
\definecolor{light-gray}{gray}{0.93}
\definecolor{lightblue}{rgb}{0.85, 0.95, 1}
\definecolor{lighterblue}{rgb}{0.9, 0.97, 1}
\definecolor{palestblue}{rgb}{0.95, 0.98, 1}

\usepackage[ruled,vlined,linesnumbered]{algorithm2e}
\definecolor{revA}{RGB}{220, 20, 60}    
\definecolor{revB}{RGB}{30, 144, 255}   
\definecolor{revC}{RGB}{34, 139, 34}    

\definecolor{codegreen}{rgb}{0,0.6,0}
\definecolor{codegray}{rgb}{0.5,0.5,0.5}
\definecolor{codepurple}{rgb}{0.58,0,0.82}
\definecolor{backcolour}{rgb}{0.95,0.95,0.92}

\lstdefinestyle{mystyle}{
    backgroundcolor=\color{backcolour},   
    commentstyle=\color{codegreen},
    keywordstyle=\color{magenta},
    numberstyle=\tiny\color{codegray},
    stringstyle=\color{codepurple},
    basicstyle=\ttfamily\footnotesize, 
    breakatwhitespace=false,         
    breaklines=true,                 
    captionpos=t,                    
    keepspaces=true,                 
    numbers=left,                    
    numbersep=5pt,                  
    showspaces=false,                
    showstringspaces=false,
    showtabs=false,                  
    tabsize=4
}

\def\paperID{33407} 
\def\confName{CVPR}
\def\confYear{2026}

\title{AdaSpark: Adaptive Sparsity for Efficient Long-Video Understanding}

\author{Handong Li\textsuperscript{1,2,3\thanks{Equal Contribution.}} \quad Zikang Liu\textsuperscript{1,2,3*} \quad Longteng Guo\textsuperscript{1,2*} \quad Tongtian Yue\textsuperscript{1,2,3} \quad Yepeng Tang\textsuperscript{2} \\
\quad Xinxin Zhu\textsuperscript{1,2} \quad Chuanyang Zheng\textsuperscript{3,4} \quad Ziming Wang\textsuperscript{3,4} \quad Zhibin Wang\textsuperscript{3,4} \\
\quad Jun Song\textsuperscript{3,4} \quad Cheng Yu\textsuperscript{3,4} \quad Bo Zheng\textsuperscript{3,4} \quad Jing Liu\textsuperscript{1,2\thanks{Corresponding Author.}} \\
\textsuperscript{1}School of Artificial Intelligence, University of Chinese Academy of Sciences \\
\textsuperscript{2}Institute of Automation, Chinese Academy of Sciences \\ 
\textsuperscript{3}Alibaba Group Holding Limited \quad \textsuperscript{4}Future Living Lab of Alibaba}

\makeatletter
\renewcommand\paragraph{\@startsection{paragraph}{4}{\z@}%
    {-0.8ex \@plus -0.2ex \@minus -0.1ex}
    {-1em}
    {\normalfont\normalsize\bfseries}}       
\makeatother

\begin{document}
\maketitle
\begin{abstract}

Processing long-form videos with Video Large Language Models (Video-LLMs) is computationally prohibitive. Current efficiency methods often compromise fine-grained perception through irreversible information disposal or inhibit long-range temporal modeling via rigid, predefined sparse patterns. This paper introduces AdaSpark, an adaptive sparsity framework designed to address these limitations. AdaSpark first partitions video inputs into 3D spatio-temporal cubes. It then employs two co-designed, context-aware components: (1) Adaptive Cube-Selective Attention (AdaS-Attn), which adaptively selects a subset of relevant video cubes to attend for each query token, and (2) Adaptive Token-Selective FFN (AdaS-FFN), which selectively processes only the most salient tokens within each cube. An entropy-based (Top-p) selection mechanism adaptively allocates computational resources based on input complexity. Experiments demonstrate that AdaSpark significantly reduces computational load by up to 57\% FLOPs while maintaining comparable performance to dense models and preserving fine-grained, long-range dependencies, as validated on challenging hour-scale video benchmarks.

\end{abstract}    
\section{Introduction}
\label{sec:intro}

Fine-grained understanding of long-form videos represents a critical frontier for Video Large Language Models (Video-LLMs) \cite{Qwen2VL,Qwen2_5VL,zhang2024video,li2025breaking}. While significant strides have been made in the domain of short-video understanding \cite{li2024mvbench,chen2023vast,chencosa,zhang2025scaling,liu2023enhancing}, scaling these successes to long-form videos presents a formidable challenge. Long videos can easily generate token sequences exceeding tens of thousands or even approaching 1M tokens. Applying standard Video-LLMs to such inputs is computationally infeasible. The core bottlenecks are twofold: the \textit{quadratic complexity ($O(N^2)$) of the self-attention mechanism} and the \textit{substantial activation costs of the Feed-Forward Neural Networks (FFNs)}. Consequently, the computational load escalates dramatically, rendering training and inference on high-resolution, long-duration videos prohibitively expensive.

In response to this scalability crisis, a plethora of efficient designs have been proposed to mitigate the computational burden \cite{yuframe,he2024ma,chen2024image,shu2025video,jin2024chat,fu2024framefusion,li2024llama,liu2025vrope}. Despite achieving commendable trade-offs between efficiency and accuracy, we observe that these methods suffer from a few limitations, which we categorize into two primary drawbacks.

(1) \textbf{Compromised Fine-Grained Perception.} Many existing strategies rely on heuristic information disposal to reduce the effective sequence length. Techniques such as \textit{frame sampling} \cite{yuframe,zhang2025q,tang2025adaptive} or \textit{token pruning} \cite{chen2024image,xing2024pyramiddrop,zhangsparsevlm} permanently discard spatio-temporal tokens, either prior to model ingress or dynamically during computation. While effective at reducing cost, this irreversible information loss risks eliminating subtle visual details or brief, yet crucial, events. Similarly, \textit{token compression} \cite{bolya2023token,he2024ma,li2024vidtome,shentempme}, which fuses features based on similarity, can corrupt or average out distinct information, leading to a loss of perceptual fidelity required for fine-grained tasks.

(2) \textbf{Inhibited Long-Range Temporal Modeling}
The second drawback stems from the imposition of rigid, pre-defined computational patterns. Approaches like \textit{local attention} \cite{li2025mminference,laiflexprefill,xuxattention} mechanisms replace the full $O(N^2)$ matrix with a restricted, computationally cheaper pattern. It explicitly breaks global sequence connectivity and restricts information flow to fixed local windows. This structural constraint fundamentally undermines the model's capacity to capture complex, long-range dependencies across extended temporal spans. Furthermore, information-discarding methods (like sampling) also contribute to this issue by creating temporal discontinuities, making it difficult to model continuous, evolving causal relationships.

\begin{figure}[t]
    \centering
    \includegraphics[width=1.0\linewidth]{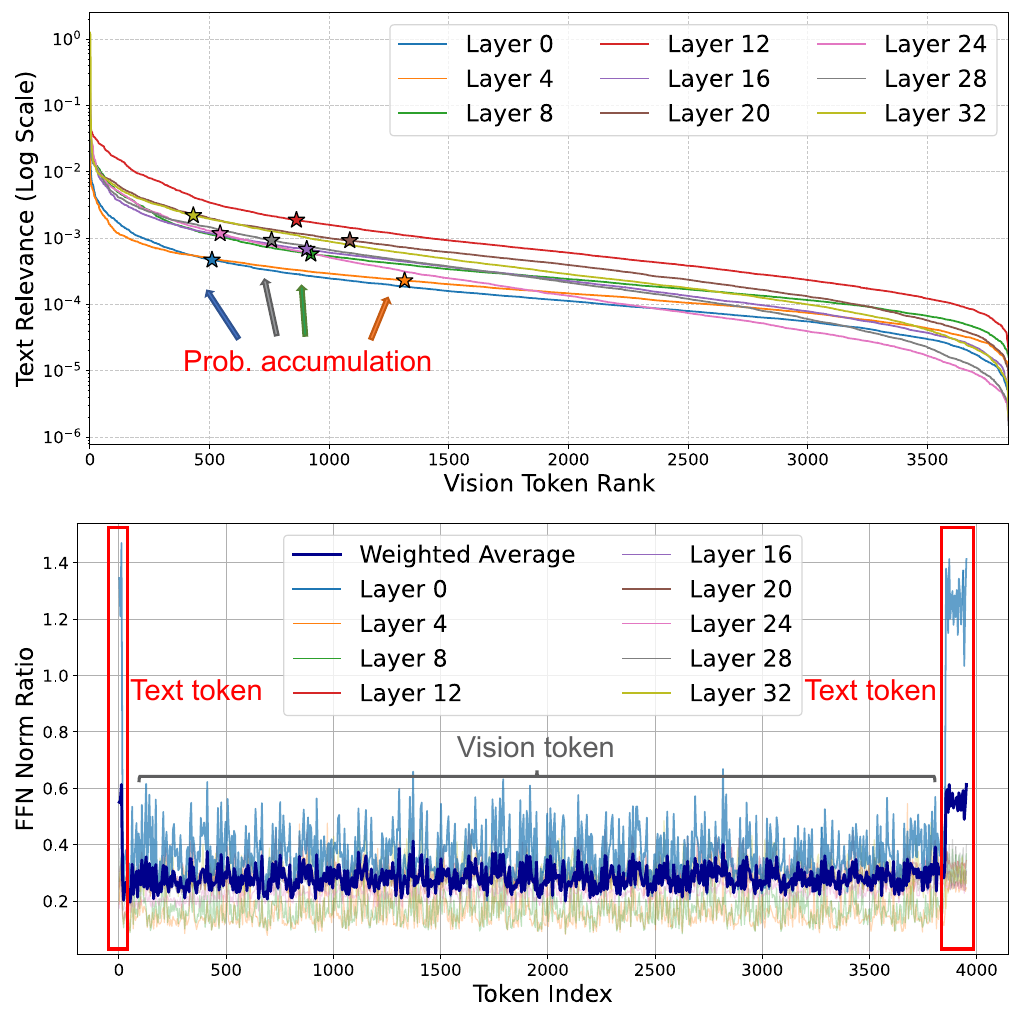}
    \vspace{-7mm}
    \caption{\textbf{Preliminary analysis.} We analyzed internal distributions within the video-LLM layers. The upper figure shows text-to-visual attention score distributions, marking the 0.7 cumulative probability point per layer with a star. The lower figure displays L2 norm changes across modalities after the FFN, quantified as the post-to-pre norm ratio.}
    \vspace{-6mm}
    \label{fig:pre_exp}
\end{figure}

To address the above challenges, we first conduct a preliminary analysis on the Qwen2.5-VL \cite{Qwen2_5VL} model, as illustrated in Figure \ref{fig:pre_exp}. Our findings are twofold:

(1) \textbf{The video attention exhibit high intrinsic sparsity.} The upper plot of Figure \ref{fig:pre_exp} shows that the attention weights are highly concentrated, with a small subset of vision tokens consistently capturing the majority of the cumulative attention probability. This confirms the inherent sparsity of the attention structure. Furthermore, the number of tokens required to reach a fixed cumulative probability threshold varies significantly across different layers. This observation strongly suggests that employing a static, fixed sparsity ratio would be suboptimal, as it would inevitably lead to information loss in layers requiring denser attention.

(2) \textbf{The FFN layers exhibit computational inertia for visual tokens.} The lower plot of Figure \ref{fig:pre_exp} reveals a stark contrast in how FFN layers process different modalities. Text tokens (highlighted by the red boxes) undergo a substantial transformation, as evidenced by the large variance and high spikes in their L2-norm ratio (post-FFN vs. pre-FFN). In sharp contrast, the vast majority of vision tokens exhibit a much more stable and consistent norm ratio. This implies that the transformative effect of the FFN on many video tokens is significantly less dynamic, providing a strong justification for a sparse design that selectively applies FFN computation.

These observations collectively indicate that a large portion of computation in long-video processing is redundant. Motivated by this, we introduce \textbf{Adaptive Cube-Token Sparsity (AdaSpark)}, which applies adaptive sparsity at both the cube level and the token level in a coordinated fashion, selecting informative 3D spatio-temporal cubes for attention while identifying salient tokens inside each cube for FFN computation, thereby reducing unnecessary computation without sacrificing fine-grained detail or long-range temporal dependencies.

Our method first leverages the inherent 3D spatio-temporal locality of video by partitioning the input into semantically related \textit{cubes}, each with a shape of $h \times w \times t$. This partitioning underpins our unified adaptive conditional computation strategy, which is implemented via two core components. The \textbf{Adaptive Cube-Selective Attention (AdaS-Attn)} mechanism introduces sparsity to self-attention; rather than computing full attention, each \textit{query} token adaptively selects a subset of video cubes to serve as the \textit{keys} and \textit{values}. On the other hand, the \textbf{Adaptive Token-Selective FFN (AdaS-FFN)} component addresses the feed-forward layers by identifying and processing only the most salient tokens within each video cube (i.e., those with the largest L2-norm). The remaining, less salient tokens bypass this expensive computation via a residual connection, significantly reducing the computational load. We further introduce an entropy-based mechanism (Top-p selection) that makes the sparsity fully context-aware, allowing the model to intelligently adapt its computational budget based on the information density of the current input.

Comprehensive experiments demonstrate that our method, AdaSpark, achieves strong performance across a suite of diverse long-video benchmarks, consistently outperforming both dense baselines and prior efficiency-focused methods. This is achieved while substantially reducing the computational burden, achieving up to a 57\% reduction in FLOPs compared to the dense backbone. In summary, our contributions can be summarized as follows:
\begin{itemize}
    \item We propose AdaSpark, a novel and efficient Video-LLM framework that introduces an adaptive unified sparse strategy to mitigate the prohibitive computational cost of processing long-video sequences.

    \item We develop two complementary, entropy-guided sparse mechanisms for video that establish hierarchical cube–token sparsity, adaptively selecting informative 3D spatio-temporal cubes and the salient tokens within them to adapt computation to input complexity.

    \item We demonstrate through extensive experiments that AdaSpark achieves strong performance on a suite of long-video benchmarks while achieving up to a 57\% reduction in FLOPs compared to the dense backbone.
\end{itemize}

\section{Related Work}

\subsection{Long Video Understanding} When video durations extend to the hour scale, visual token sequences can approach 300,000, introducing prohibitive computational complexity. Some works~\cite{shang2025llava,zhong2025aim,uzkent2023dynamic,ye2025fit,lu2025vipe,zhang2025learning,tangdivid} aim to reduce visual tokens before core model ingestion, typically achieved through compression, merging, or downsampling techniques. For instance, LLaMA-VID~\cite{li2024llama} and Video-CCAM~\cite{fei2024video} utilize cross-attention mechanisms to compress visual features into fixed-length query embeddings. Furthermore, VideoChat-Flash~\cite{li2024videochat} directly merges similar visual tokens via the ToMe~\cite{bolya2023token} methodology, whereas VideoLLaMA2~\cite{cheng2024videollama} employs spatio-temporal convolutions for effective visual token downsampling. FrameFusion~\cite{fu2024framefusion} achieves more efficient context compression by implementing merging and pruning operations based on inter-frame attention scores.

Inspired by advancements in long-context LLMs, it is evident that exhaustive, full-attention computation between all query and key vectors is frequently unnecessary. Building on this observation, MMInference~\cite{li2025mminference} introduced pre-defined sparse patterns to adaptively optimize the computational FLOPs for distinct attention heads, thereby achieving notable optimization during the pre-filling stage. VideoChat-Flash~\cite{li2024videochat} attempts to accelerate both pre-filling and decoding by eliminating KVs in specific layers that exhibit low relevance to the query text. Video-XL~\cite{shu2025video} partitions the video input into chunks and processes them streamingly, generating sparse KVs from special tokens inserted among the visual tokens. While these works explore information redundancy in long video understanding, they predominantly rely on fixed sparsity paradigms or lack a unified design across multiple modules.

\subsection{Sparse Network Design} 

Recently, significant advancements have emerged in long-context sparse modeling for LLMs. As an alternative to conventional approaches, such as structural constraints based on attention sinks, dynamic inference-time sparsity, or linear-like models requiring training from scratch, a promising paradigm known as Native Sparse Modeling has appeared. This approach functions by allowing the model to determine where to attend without relying on predefined biases. Such an architecture facilitates a seamless transition between full and sparse attention modes, thereby maximizing compatibility with existing pre-trained models and enabling both efficient inference and accelerated training without compromising performance. This direction has spurred the development of methods like NSA~\cite{yuan2025native}, whose 'select' branch partitions the context into blocks and computes attention on a top-k selection based on attention scores. Similarly, MoBA~\cite{lu2025moba} divides the context into chunks and selects the top-k chunks based on inter-chunk similarity. 
However, such trainable sparse modeling approaches are incapable of adaptively selecting the number of chunks for long input contexts and necessitate specific architectural designs tailored for long video.
\section{Method}~\label{sec:method}
\begin{figure*}[ht]
    \centering
    \includegraphics[width=1.0\linewidth]{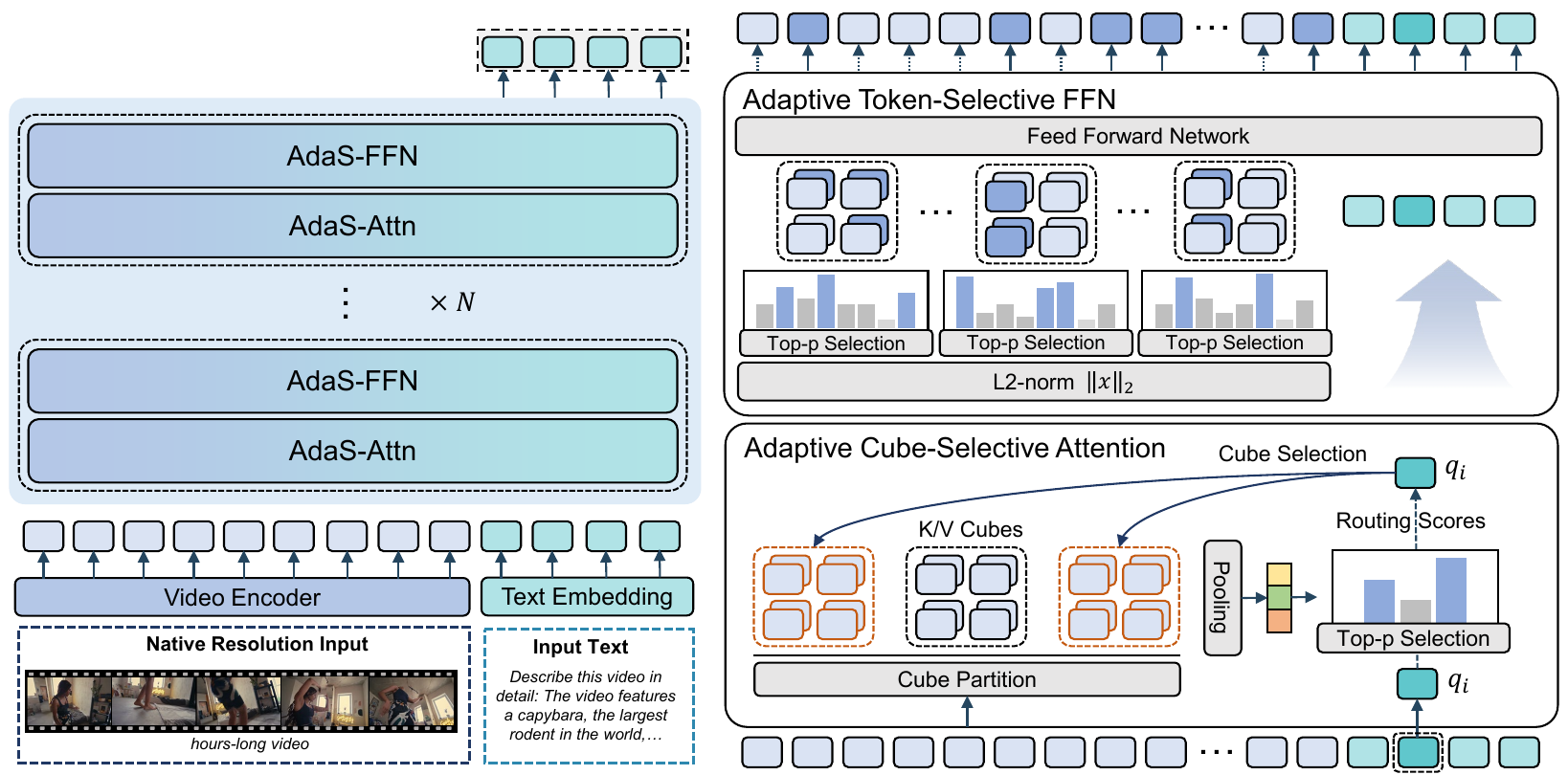}
    \vspace{-5mm}
    \caption{\textbf{Framework illustration of AdaSpark.} We process long-duration videos at their native resolution and subsequently apply video cube partitioning. Within the AdaS-Attn layer, each token query performs adaptive selection based on relevance scores computed over preceding Cubes. Upon entering the AdaS-FFN, visual tokens within each Cube are adaptively selected to pass through the FFN, while the transformations for the remaining tokens are estimated via Mean Compensation. Text tokens pass densely through the AdaS-FFN.}
    \vspace{-5mm}
    \label{fig:framework}
\end{figure*}

Motivated by the high spatio-temporal redundancy inherent in video data and the intrinsic sparsity of LLMs, we propose AdaSpark, a unified and efficient sparse strategy applied to both Attention and FFN components (Figure~\ref{fig:framework}), to reduce the computational costs of model inference.

\subsection{Video Cube Partition}
AdaSpark applies sparse computation at the granularity of \textit{Cube-Token} levels. This design choice mandates that tokens encapsulated within a single cube should be as semantically homogeneous as possible (i.e., possess high semantic cohesion). This cohesion is critical for enhancing the accuracy and stability of the sparse selection mechanism.

Unlike one-dimensional natural language, video data is inherently characterized by a 3D spatio-temporal structure, exhibiting strong locality; proximate tokens in this 3D space are highly likely to be correlated. We therefore leverage this intrinsic property by partitioning the video tokens fed into the LLM according to a cube window with a shape of $h \times w \times t$. These resulting \textit{cubes} serve as the atomic units for our subsequent sparse attention and FFN algorithms.

\subsection{Adaptive Cube-Selective Attention}
\label{sec:sparse_attention}

In the attention component, we introduce an Adaptive Cube-Selective Attention (AdaS-Attn) for video tokens, computed per attention head. For any given video query token, $q$, belonging to the $i$-th cube $C_i$ (where $i \in \{1, 2, \dots, N\}$ and $N$ is the total number of cubes), we first determine its relevance to all \textit{preceding} cubes. This is achieved by computing a selection similarity score between $q$ and the \textbf{mean key vector} ($\bar{k}_j$) of each preceding block $C_j$ (where $j < i$).

This process yields a probability distribution $P_i$ over the preceding blocks:
\begin{equation}
P_i = \text{Softmax}\left( \left[ \frac{q \cdot \bar{k}_1}{\sqrt{d_k}}, \frac{q \cdot \bar{k}_2}{\sqrt{d_k}}, \dots, \frac{q \cdot \bar{k}_{i-1}}{\sqrt{d_k}} \right]^T \right)
\label{eq:selection_similarity}
\end{equation}
where $d_k$ is the dimension of the key vectors and $\bar{k}_j = \text{Mean}(k \in K_j)$ is the average key vector for all tokens in cube $C_j$.

Next, we employ an entropy-based, adaptive selection strategy to determine the set of cubes to attend to. We utilize \textbf{nucleus (Top-p) sampling} on the distribution $P_i$. This strategy is inherently context-aware:
\begin{equation}
\mathcal{S}_i = \{ j \mid j < i \text{ and } p_j \in \text{Top-p}(P_i, p) \}
\label{eq:nucleus_selection}
\end{equation}
where $\mathcal{S}_i$ is the selected set of block indices and $p_j$ is the $j$-th probability from $P_i$ (Eq. \ref{eq:selection_similarity}). This approach adaptively adapts the sparsity level. For a high-entropy (flat) distribution, where similarity is dispersed, more cubes are selected to aggregate sufficient information. Conversely, for a low-entropy (sharp) distribution, where relevance is concentrated, only the few most pertinent cubes are chosen.

Finally, based on the proximity hypothesis---positing that adjacent tokens are most critical---the query $q$ \textit{always} performs full attention on the preceding tokens within its \textit{own} cube (denoted $K_i^{<q}$ and $V_i^{<q}$). The final video token attention output is computed by attending to the concatenation of the selected sparse cubes and the local cube:
\begin{equation}
\begin{aligned}
K_{\text{att}} &= \text{Concat}\left( \left[K_j \text{ for } j \in \mathcal{S}_i\right], K_i^{<q} \right) \\
V_{\text{att}} &= \text{Concat}\left( \left[V_j \text{ for } j \in \mathcal{S}_i\right], V_i^{<q} \right) \\
\text{Out}(q) &= \text{Softmax}\left(\frac{q K_{\text{att}}^T}{\sqrt{d_k}}\right) V_{\text{att}}
\end{aligned}
\end{equation}
where $K_j$ and $V_j$ represent all keys and values in cube $C_j$.
For text tokens, cube-wise attention is applied to preceding visual tokens, while standard token-wise attention is applied to text tokens.

\subsection{Adaptive Token-Selective FFN}
\label{sec:sparse_mlp}
Leveraging the high semantic cohesion within each video cube $C_i$, our Adaptive Token-Selective FFN (AdaS-FFN) efficiently processes only the most salient tokens through the expensive FFN transformations, while the remaining tokens bypass this computation. We hypothesize that tokens which are more information-rich exhibit a higher L2-norm (${\|x\|_2}$)~\cite{abbasi2025normxlogit}.

Specifically, for all token embeddings $x_j$ within a given cube $C_i$, we first compute an importance score $s_j$ by L1-normalizing their L2-norms. This creates a probability distribution $S_i$ representing the relative importance of each token in the cube:
\begin{equation}
s_j = \frac{\|x_j\|_2}{\sum_{k \in C_i} \|x_k\|_2 + \epsilon}
\label{eq:importance_score}
\end{equation}
where $\epsilon$ is a small constant for numerical stability.

Analogous to our attention strategy, we employ an entropy-aware nucleus (Top-p) selection on the importance distribution $S_i$ to identify the set of tokens to activate, $\mathcal{M}_i$:
\begin{equation}
\mathcal{M}_i = \{ j \mid j \in C_i \text{ and } s_j \in \text{Top-p}(S_i, p) \}
\label{eq:mlp_selection}    
\end{equation}
This adaptively selects only the most prominent tokens for full FFN computation. Let $\mathcal{R}_i = C_i \setminus \mathcal{M}_i$ denote the set of remaining (unselected) tokens.

The final computation is then bifurcated. Tokens in the activated set $\mathcal{M}_i$ are processed by the full FFN (which includes its own residual connection):
\begin{equation}
y_j = x_j + \text{FFN}(x_j), \quad \forall j \in \mathcal{M}_i
\end{equation}

The remaining tokens $x_k$ in $\mathcal{R}_i$ bypass the expensive FFN. 
We apply a \textbf{Mean Compensation }strategy for those skipped tokens. We first compute the activated set's mean transformation $\overline{m}_{i}$, then use it as an estimate of those remaining tokens' transformation, and apply their own residual connection:
\begin{equation}
\begin{aligned}
y_k &= x_k + \bar{m}_i, \quad \forall k \in \mathcal{R}_i, \\
\bar{m}_i &= \frac{1}{|\mathcal{M}_i|} \sum_{j \in \mathcal{M}_i} \text{FFN}(x_j)
\end{aligned}
\end{equation}
This strategy significantly reduces computation while ensuring all tokens are updated.
Finally, similar to the attention layer, all text tokens $x_t$ undergo the full, dense FFN computation to preserve their rich instructional and semantic content.

In summary, our AdaS-Attn and AdaS-FFN provide a coherent framework for efficient long-video modeling. By leveraging semantically cohesive cubes as the fundamental computational unit, we apply context-aware, entropy-based sparsity to \textit{both} the attention and FFN. This holistic approach significantly reduces computational and memory overhead, while preserving the critical fine-grained information and global long-range dependencies necessary for high-fidelity long video understanding.

\section{Experiment}~\label{sec:exp}
In this section, we present the experimental setup, details of the datasets, and comparisons with state-of-the-art efficient methods across multiple video benchmarks

\begin{table*}[t]
\centering
\captionof{table}{\textbf{Performance on zero-shot video-language benchmarks.} We evaluate  AdaSpark on 1 extra long video benchmarks, 4 long-form video benchmarks, 1 short video benchmark, 1 spatial perception video benchmark and 1 video grounding benchmark. The results in \textbf{bold} and \underline{underline} values indicate the best and second-best performance among encoder-free models, respectively.}
\vspace{-2mm}
\label{tab:mllm}
\resizebox{1.0\linewidth}{!}{
\begin{tabular}{l c ccc ccccc}
\toprule
\multirow{2}{*}{\textbf{Model}}  & \multicolumn{1}{c}{\textbf{Extra Long}}  & \multicolumn{4}{c}{\textbf{Long-form}} & \multicolumn{1}{c}{\textbf{Short-form}} & \multicolumn{1}{c}{\textbf{Spatial}} & \multicolumn{1}{c}{\textbf{Grounding}}\\ 
\cmidrule(lr){2-2} \cmidrule(lr){3-6} \cmidrule(lr){7-7} \cmidrule(lr){8-8} \cmidrule(lr){9-9} 
&   \multirow{1}{*}{\textbf{VideoNIAH}} &   \multirow{1}{*}{\textbf{MLVU Dev}} & \multirow{1}{*}{\textbf{VideoMME}} & \multirow{1}{*}{\textbf{LongVideo}}  & \multirow{1}{*}{\textbf{LVBench}}& \multirow{1}{*}{\textbf{MVBench}}& \multirow{1}{*}{\textbf{VsiBench}}& \multirow{1}{*}{\textbf{CharadesSTA}}\\ 
\midrule
\multicolumn{1}{l}{\textbf{\textit{Small Size Models}}} \\
LongVU-3B~\cite{shenlongvu} & - & 55.9   & 51.5 &  -  &  -  &  60.9  &  -  &  - \\
InternVL2.5-2B~\cite{chen2024expanding} & - & 61.4 & 51.9  &  52.0  &  -  &  \underline{68.8}  &  -  &  - \\
VideoChat-Flash-2B~\cite{li2024videochat}& \underline{92.0} & \underline{65.7}  & 57.0  & \textbf{58.3}  & 42.9   &   \textbf{70.0} & -   &  \textbf{45.2} \\
Qwen-2.5-VL-3B~\cite{Qwen2_5VL}& 86.5 & 65.3   & 61.5 & 54.2  & {43.3}  & 65.7   & 32.4 & 42.6 \\
Qwen-2.5-VL-3B~\cite{Qwen2_5VL} + SFT  & 88.0 &  65.2   &  60.3   &   53.6   &  43.1   &   64.9  &   33.1  &  43.6  \\
\rowcolor{palestblue} ~ + FastV~\cite{chen2024image} & 88.0 & 65.1   & 62.4 & 55.9  & 43.3  & 65.6  & 29.9  & 41.1 \\
\rowcolor{palestblue} ~ + ToMe~\cite{bolya2023token} & 86.5 & 65.4  & 62.2 & 55.8  & 42.4  & 64.5  & 29.4  & 43.3 \\
\rowcolor{palestblue} ~ + MoBA~\cite{lu2025moba} & 65.5 & 63.2   & 58.4 & 51.2 & 39.4  & 64.0  & 28.9   & 40.1 \\
\rowcolor{palestblue} ~ + FrameFusion~\cite{fu2024framefusion} & 90.0 & 65.0 & \underline{62.4} & 56.0  & \underline{43.4}  & 65.4  & \underline{33.6}  & 41.4 \\
\rowcolor{lightblue} \textbf{~ + AdaSpark~(Ours)} & \textbf{95.5} & \textbf{67.3}   &  \textbf{63.5}  &  \underline{56.3}   &  \textbf{45.0}  &  {66.5}  &  \textbf{35.8}  &  \underline{45.0} \\
\midrule
\multicolumn{1}{l}{\textbf{\textit{Mid Size Models}}} \\
LLaMA-VID-7B~\cite{li2024llama} & - & 33.2  & -  &  -  &  - &   41.4 &  -  &  - \\
VideoChat2-7B~\cite{li2023mvbench} & - & 47.9  & 39.5  &  39.3  &  - &   62.3 &  -  &  - \\
LongVA-7B~\cite{zhang2024long} & 58.0 & 56.3  & 52.6  &  47.8  &  - &   - &  -  &  - \\
Video-XL-7B~\cite{shu2025video} & 90.0 & 64.9  & 55.5  &  50.7  &  - &   55.3 &  -  &  - \\
Qwen-2.5-VL-7B~\cite{Qwen2_5VL}& 83.0 & \underline{68.3}   & 65.1 & 60.7  & \underline{45.3}  & \underline{69.6}   & 35.6  & 52.4 \\
Qwen-2.5-VL-7B~\cite{Qwen2_5VL} + SFT  & 86.0 &  68.1    &  64.7   &   59.8   &  45.2   &  67.0   &   36.7  &  \underline{53.9}  \\
\rowcolor{palestblue} ~ + FastV~\cite{chen2024image} & 88.5 & 65.7   & 66.0 & 60.8  & 43.3  & 67.0  & 39.2  & 51.3 \\
\rowcolor{palestblue} ~ + ToMe~\cite{bolya2023token} & 85.0 & 66.1  & 65.8 & 60.8  & 43.1  & 67.2  & 38.7  & {53.5} \\
\rowcolor{palestblue} ~ + MoBA~\cite{lu2025moba} & 70.5 & 64.7  & 63.4 & 58.3  & 42.3  & 67.5  & 34.8  & 50.7 \\
\rowcolor{palestblue} ~ + FrameFusion~\cite{fu2024framefusion}& \underline{92.5} & 66.3  & \underline{66.1} & \underline{61.1}   & 42.8  & 66.8  & \underline{39.3}  & 51.3 \\
\rowcolor{lightblue} \textbf{~ + AdaSpark~(Ours)} & \textbf{97.5} & \textbf{69.8}    & \textbf{66.2}  &  \textbf{62.1}   &  \textbf{47.9}  &  \textbf{70.3}  &  \textbf{39.8}  &  \textbf{55.3} \\
\bottomrule
\end{tabular}
}
\end{table*}

\paragraph{Implementation Details.} 
We apply our AdaSpark method on the Qwen2.5-VL-3B~\cite{Qwen2VL} and Qwen2.5-VL-7B backbones, thereby accommodating distinct model scales. Video inputs are sampled at 4 frames per second (fps) to preserve comprehensive temporal information while maintaining native resolution for spatial perception. This configuration supports a maximum visual sequence length of 48k tokens and a maximum context length of 64k tokens, the latter being double that of the backbone architecture. Subsequently, a $8\times8\times4$ spatio-temporal window is employed to partition the visual tokens into blocks. A top-p sampling threshold of $p=0.7$ is applied within both the sparse attention mechanism and the sparse Feed-Forward Network (FFN) component. Throughout training, the visual encoder remains frozen. The model is trained using a learning rate of $2\times10^{-6}$ and a global batch size of 256. The entire post-training procedure completes in approximately 4 days utilizing 32 NVIDIA H100 GPUs.

\paragraph{Training Data.} 
Our training methodology incorporates a mixed dataset, centrally featuring the llava-video-178k~\cite{zhang2024video} dataset, which serves as a foundational corpus for basic video understanding. In addition, we augment this data with 77k timestamp-grounded samples from DideMo~\cite{anne2017localizing} and ActivityNet Captions~\cite{krishna2017dense} to enhance the model's capacity for identifying key temporal information. Additional hyper-parameter details are available in the Appendix.

\subsection{Main Results}~\label{sec:exp:mllm}
We evaluate AdaSpark on a series of comprehensive video-language benchmarks: 1) Extra Long Video Understanding, using Video Needle in a Haystack~\cite{zhaoneedle,zhang2024long}; 2) Long Video Understanding, which includes MLVU~\cite{zhou2024mlvu}, VideoMME~\cite{fu2024video}, LongVideoBench~\cite{wu2024longvideobench}, and LVBench~\cite{wang2025lvbench}; 3) Short Video Understanding, using MVBench~\cite{li2023mvbench}; 4) Spatial Reasoning, with VSIBench~\cite{yang2025thinking}; and 5) Video Grounding, utilizing CharadesSTA~\cite{gao2017tall}. We evaluated the AdaSpark framework against leading efficiency-focused methods by integrating them into the Qwen-2.5-VL-3B and Qwen-2.5-VL-7B backbones. For these competing methods, we utilized the optimal settings as provided in their original publications. For MoBA, we employed the same post-training data and number of training steps as our method to ensure a fair comparison. We utilized a unified evaluation script from lmms-eval~\cite{zhang2024lmmsevalrealitycheckevaluation,lmms_eval2024} and re-evaluated the performance of Qwen-2.5-VL to facilitate a fair comparison.

As detailed in Table \ref{tab:mllm}, our method demonstrated superior performance across key capability axes. On the extra-long video benchmark, Video-NIAH, AdaSpark outperforms all sparse methods and classical long-video models, indicating our method's efficient extraction of key information (detailed experimental comparisons and visualizations are provided in Section~\ref{extralong}). In Long Video Understanding, AdaSpark significantly outperformed other efficient methods and the backbone model; our 7B model achieved top scores on all four benchmarks, while the 3B model led on three. This validates that our adaptive sparse methodology effectively retains long-temporal information while reducing computational cost. In Spatial Reasoning, AdaSpark matched the strongest sparse attention baselines and surpassed other efficient methods, confirming its preservation of spatial fidelity. Our model also surpassed all efficient methods in Short Video Understanding, with distinct advantages in Video Grounding (3B: 45.0, 7B: 55.3).

Conversely, MoBA exhibited the weakest performance, which we attribute to its failure to perform meaningful visual sequence partitioning. In contrast, AdaSpark's superior performance on several benchmarks is ascribed to its effective spatio-temporal feature preservation. These comprehensive results validate that our adaptive sparse design consistently yields significant performance gains, enabling efficient, high-fidelity video understanding.

\subsection{Extra Long Video Evaluation}
\label{extralong}
\begin{figure*}[ht]
    \centering
    \includegraphics[width=1.0\linewidth]{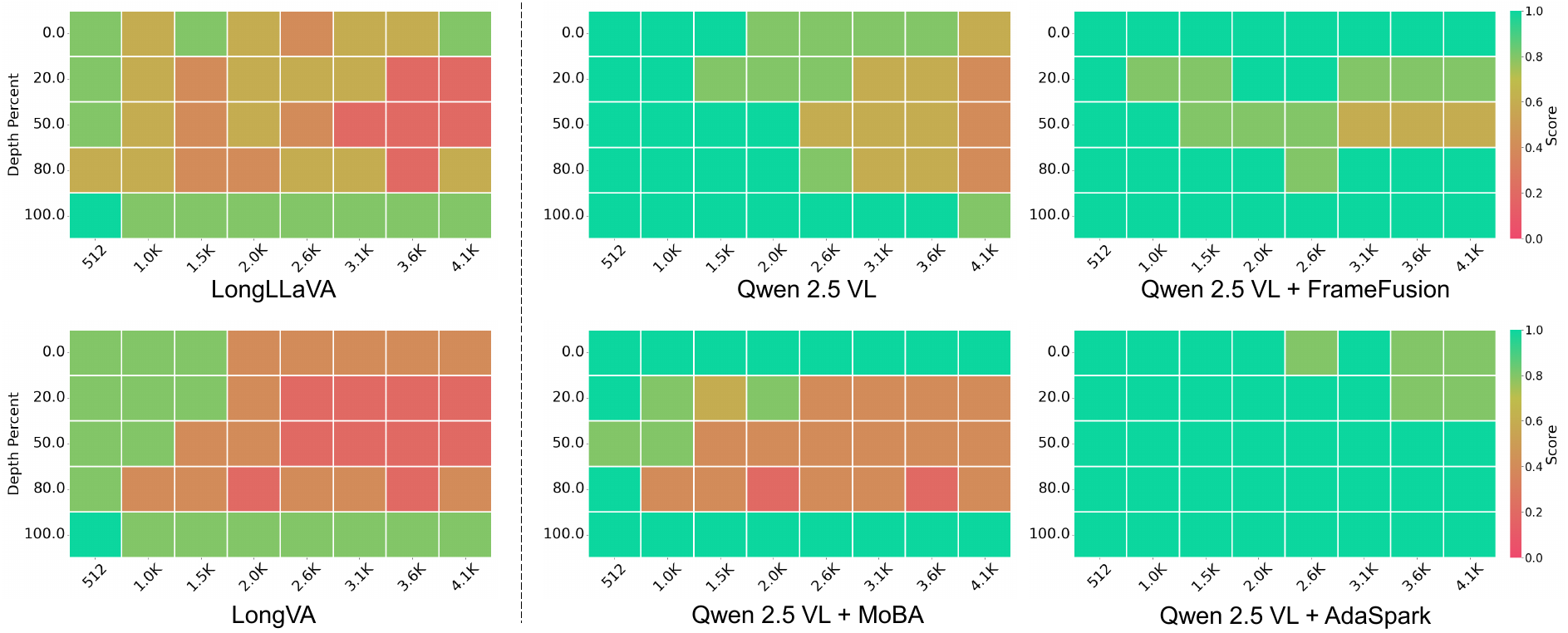}
    \vspace{-7mm}
    \caption{\textbf{Video Needle in A Haystack Results.} We compare AdaSpark against existing high-efficiency models and methods.}
    \vspace{-3mm}
    \label{fig:niah}
\end{figure*}

To further assess our model's performance on extended-duration video inputs, we conducted a video ``Needle in a Haystack" (VideoNIAH) evaluation. This test is designed to measure the model's ability to retrieve specific, localized information embedded at various temporal depths within a protracted video context. As illustrated in Figure~\ref{fig:niah}, we benchmarked AdaSpark against two classical long-video models~\cite{zhang2024long,wang2024longllava} and two alternative sparse methodologies. In contrast to prior works that employed perplexity-based metrics—which we observed can be easily satisfied even with non-specific answers—we adopted a more rigorous generative NIAH evaluation. For the classical long-video models, we report the performance cited in their original publications. For the sparse methodologies, we re-implemented them uniformly using the Qwen2.5-VL backbone, processing 4,096 frames at 224 resolution, which resulted in a 300k-token context. The resulting heatmaps depict retrieval accuracy, revealing that baseline methods suffer from severe performance degradation (indicated by yellow/red patches) at various temporal depths. In sharp contrast, our model maintains consistently high retrieval accuracy across all evaluated depths and video lengths. These results confirm that AdaSpark scales effectively to extra-long video inputs, demonstrating a clear advantage by achieving superior retrieval accuracy while operating under a minimal computational budget.

\begin{table*}[h]
 \centering
 \caption{\textbf{Effects of AdaS-Attn and AdaS-FFN}.}
 \label{tab:module_ablation}
 \vspace{-3mm}
 \resizebox{0.9\linewidth}{!}{
 \begin{tabular}{cccccccc}
  \toprule
  \multicolumn{2}{c}{\textbf{Module}} & \multirow{2}{*}{\textbf{TFLOPs$\downarrow$}} & \textbf{MLVU} & \textbf{VideoMME} & \textbf{LongVideo} & \textbf{LVBench} & \textbf{Charades STA}\\
   \cmidrule(lr){1-2} \cmidrule(lr){4-8}
  AdaS-Attn & AdaS-FFN &   & M-avg &  w/o sub & acc & acc & mIoU \\
  \midrule
  \ding{55} & \ding{55} & 299.5 & 65.2 &  60.3 & {53.6} & {43.1} & 43.6 \\
   \ding{55} & \textbf{\checkmark} & 216.9\textcolor{blue}{(-28\%)} & {63.9} &  \textbf{62.8} & \textbf{55.6} & {44.1} & 41.7 \\
   \textbf{\checkmark} &  \ding{55} & 213.1\textcolor{blue}{(-29\%)}  & \textbf{66.0} &  {61.2} & {54.5} & {43.4} & 43.7 \\
  \midrule
  \rowcolor{lightblue}
   \textbf{\checkmark} & \textbf{\checkmark} & 128.5\textcolor{blue}{(-57\%)}  & 65.4 &  62.1 & {55.3} & \textbf{44.3} & \textbf{44.2} \\
  \bottomrule
\end{tabular}}
 \vspace{-3mm}
\end{table*}
\begin{table}[t]
\caption{\textbf{Ablation on design choices of AdaSpark}.}
\vspace{-3mm}
\label{tab:ablation}
\centering
\resizebox{1.0\linewidth}{!}{
\begin{tabular}{ll cccc}
\toprule
{Model} & {Factors}  & Charades STA & VideoMME & MLVU \\
\midrule

\multicolumn{2}{l}{\textbf{\textit{Cube Shape}}} \\
(a) & $1 \times 1 \times 64$  & 35.3 & 52.7 & 51.3 \\
\rowcolor{lightblue}
(b) & $4 \times 4 \times 4$  & \textbf{35.5} & {54.0} & \textbf{52.1} \\
(c) & $8 \times 8 \times 1$ & 34.5 & \textbf{54.2} & 49.4 \\



\midrule
\multicolumn{2}{l}{\textbf{\textit{Cube Size}}} \\
(d) & 64   & 35.5 & 54.0 & 52.1  \\
(e) & 128 & 36.0 & {53.9} & 55.8  \\
\rowcolor{lightblue}
(f) & 256 & \textbf{36.1} & \textbf{55.7} & \textbf{57.5}  \\
(g) & 512  & {34.9} & 55.6 & {56.9}  \\
(h) & 1024  & {34.6} & 54.9 & {57.3}  \\

\midrule
\multicolumn{2}{l}{\textbf{\textit{Mean Compensation}}} \\
(i) & w/o mean & 35.0 & 54.1 & 56.4 \\
\rowcolor{lightblue}
(j) & w/ mean & \textbf{36.1} & \textbf{55.7} & \textbf{57.5} \\



\midrule

\end{tabular}
}
\vspace{-5mm}
\end{table}

\subsection{Ablation Study}~\label{sec:exp:ablate}
For the default configuration in our ablation studies, we employ our 3B model with the ViT parameters remaining frozen. To mitigate computational cost, the visual context length is constrained to 24k and sampled at 2 frames per second (fps), utilizing a spatial resolution of $224\times224$, which collectively results in 64 tokens per frame.

\paragraph{Effect of AdaS-Attn and AdaS-FFN.} We conducted an ablation study to evaluate our proposed sparse modules (Table \ref{tab:module_ablation}), using Qwen2.5VL-3B as the baseline. TFLOPs were measured via the DeepSpeed profiler~\cite{rasley2020deepspeed}. The Cube FFN module reduces computational cost (299.5 $\to$ 216.9 TFLOPs) and improves performance on VideoMME, LongVideo, and LVBench, but shows degradation on temporally sensitive tasks (MLVU, Charades-STA), suggesting a prioritization of spatial features. In contrast, the Cube Attn module improves performance across all five benchmarks, with the largest gains on MLVU and Charades-STA, while similarly reducing TFLOPs (299.5 $\to$ 213.1). The two components are complementary; when combined, they achieve superior performance to the baseline while reducing total computational cost by 57\%.

\begin{figure*}[t]
  \centering
  \begin{subfigure}[b]{0.66\linewidth}
    \includegraphics[width=\linewidth]{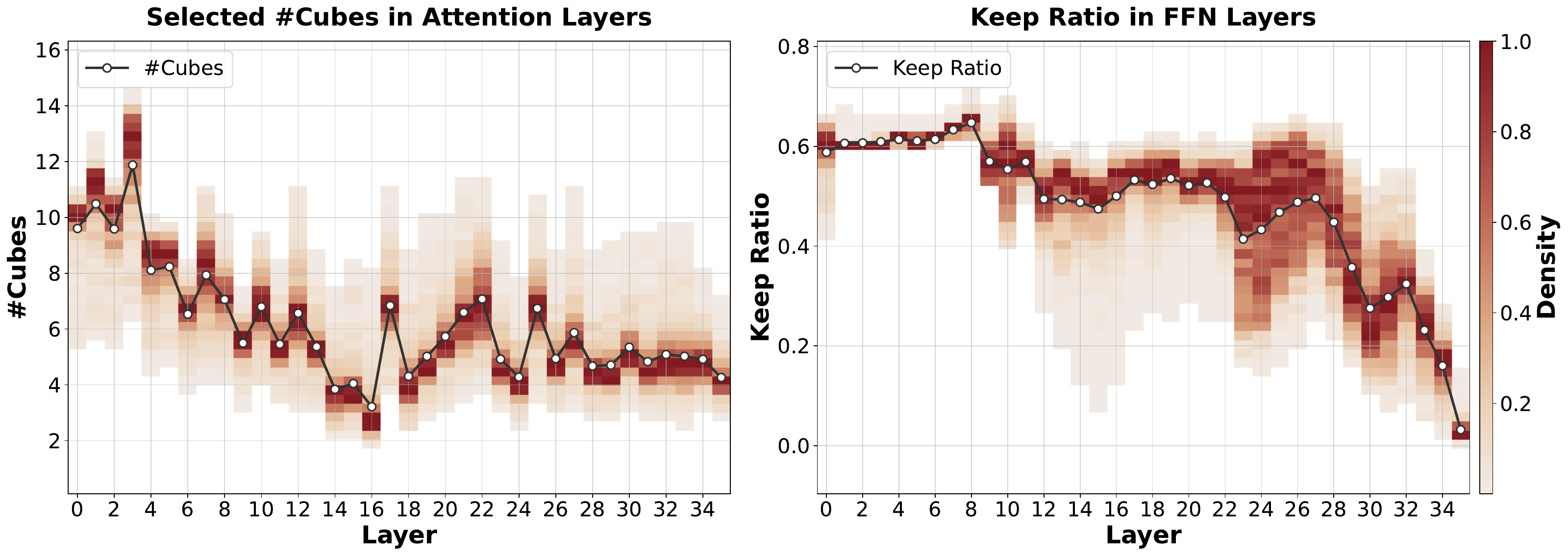}
  \end{subfigure}
  \hfill 
  \begin{subfigure}[b]{0.33\linewidth}
    \includegraphics[width=\linewidth]{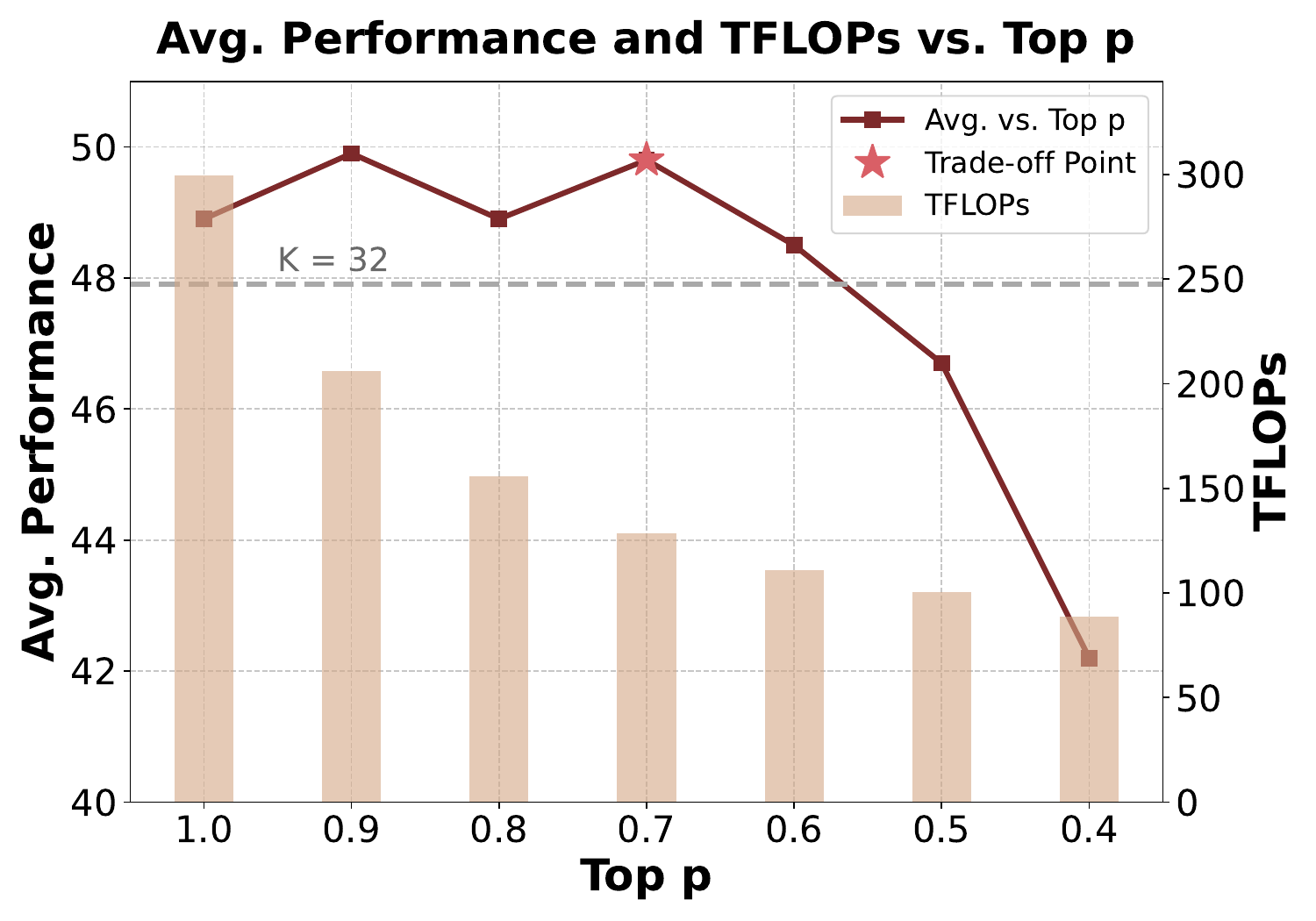}
  \end{subfigure}
  \vspace{-3mm}
  \caption{\textbf{Analysis of adaptive selection within AdaSpark.} The left figure illustrates the number of cubes selected by each query per layer in AdaS-Attn. The middle figure details the average token keep ratio per cube for each layer in AdaS-FFN. The right figure demonstrates the impact of parameter choices on the dynamic selection mechanism.}
  \label{fig:topp_layeranaly} 
  \vspace{-4mm}
\end{figure*}

\paragraph{Effect of Cube Shape.} Constraining the blocksize to 64, we subsequently investigate three different cube partitioning strategies: (a) a temporally-focused shape ($1 \times 1 \times 64$), (b) a balanced spatio-temporal shape ($4 \times 4 \times 4$), and (c) a spatially-focused shape ($8 \times 8 \times 1$). As detailed in Table \ref{tab:ablation}, the balanced $4 \times 4 \times 4$ configuration (b) achieves the most robust performance, yielding the highest scores of 35.5 on Charades STA and 52.1 on MLVU, while remaining competitive on VideoMME with a score of 54.0. This finding suggests that balancing spatial and temporal granularity yields an optimal configuration.

\paragraph{Effect of Cube Size.} We ablate the size of the token blocks, ranging from 64 to 1024. As shown in Table \ref{tab:ablation}, performance peaks with a cube size of 256 (f). This configuration achieves the highest scores across all three benchmarks, yielding 36.1 on Charades STA, 55.7 on VideoMME, and 57.5 on MLVU.

\paragraph{Effect of Mean Compensation for AdaS-FFN.} We test the effect of adding the mean transformation of activated tokens to the bypassed tokens in the sparse FFN. As shown in Table \ref{tab:ablation}, the model \texttt{w/ mean} (j) shows superior results, achieving 36.1 on Charades STA, 55.7 on VideoMME, and 57.5 on MLVU. This outperforms the model \texttt{w/o mean} (i), which scored 35.0, 54.1, and 56.4 on the respective benchmarks. This demonstrates that the computed mean feature serves as a good estimation for the tokens that bypass the MLP.

\paragraph{Effect of Top-P Adaptive Selection.} We conducted comprehensive experiments on various probability thresholds for \texttt{Top-P} selection. We compared a fixed \texttt{Top-K} selection strategy, calibrated to have an equivalent average sparsity level, against our entropy-based adaptive \texttt{Top-P} selection. In Figure~\ref{fig:topp_layeranaly} (right), we represent model performance using the average score across three benchmarks (Charades-STA, MLVU, and Video-MME), while the bar chart illustrates the computational complexity associated with each \texttt{Top-P} setting. The \texttt{Top-P} strategy significantly outperforms the \texttt{Top-K} approach when an appropriate parameter is chosen. It is observable that setting $p=0.7$ yields the best trade-off between performance and computational cost, which we adopted as our default setting. This highlights the advantage of entropy-based, content-adaptive selection.

\begin{figure}[ht]
    \centering
    \includegraphics[width=1.0\linewidth]{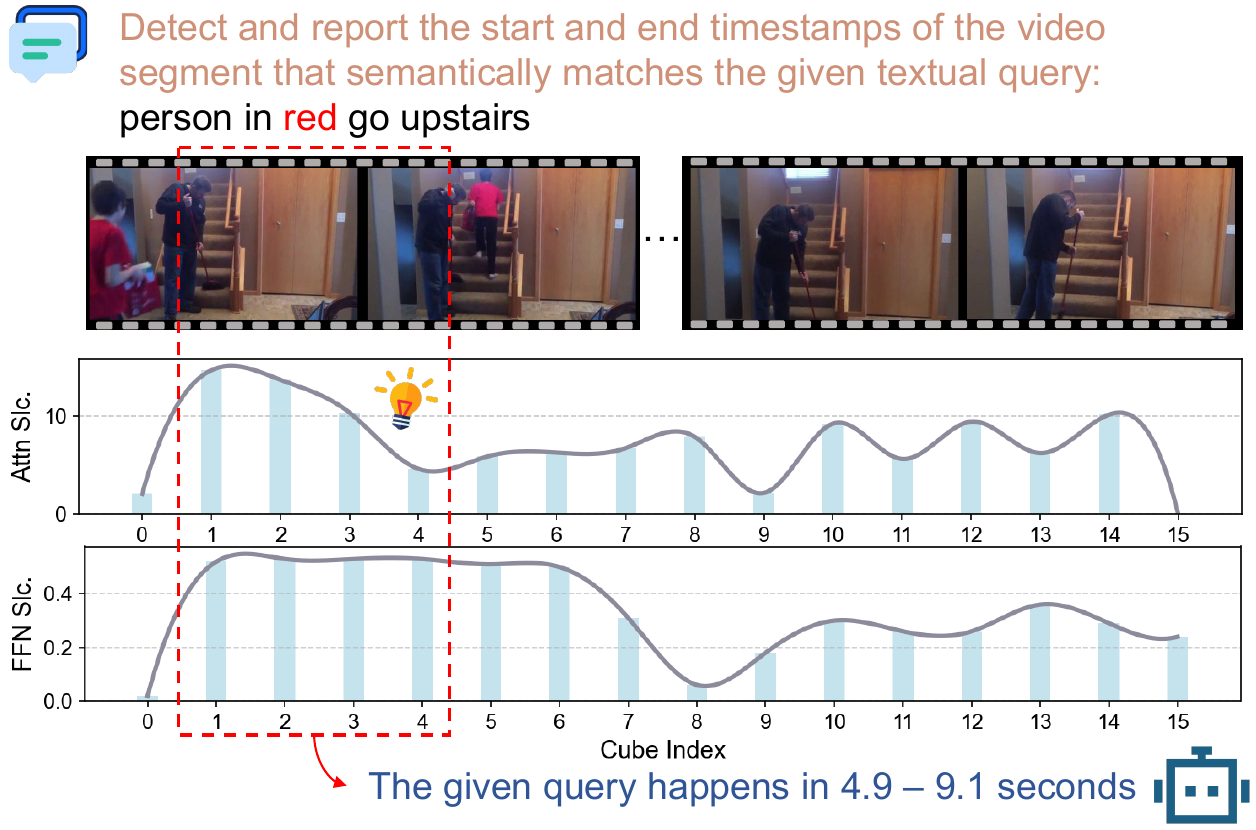}
    \vspace{-7mm}
    \caption{\textbf{Illustration of adaptive selection in a case study.} AdaSpark adaptively selects visual cubes that exhibit high relevance to the posed query token.}
    \vspace{-5mm}
    \label{fig:example}
\end{figure}

\subsection{Visualization of Adaptive Selection}

To validate the efficacy of the adaptive top-p selection mechanism, we visualized its layer-wise operation within our 3B model. Our initial analysis focused on the AdaS-Attn module. Utilizing 1,000 samples from LongVideoBench, we computed the average number of cubes selected by each query token per layer. As illustrated in Figure~\ref{fig:topp_layeranaly} (left), a distinct trend emerges: the model selects a larger quota of cubes for attention in shallower layers, progressively reducing this allocation in intermediate and deeper layers. This suggests the top-p mechanism excels at modeling foundational visual features at the outset and subsequently conserves computation by pruning attention on features that have already acquired high-level semantic representations.

Concurrently, we define a ``keep ratio" as the average sparsity granularity within each AdaS-FFN cube. As shown in Figure~\ref{fig:topp_layeranaly} (mid), this keep ratio exhibits significant layer-wise variation. Specifically, the module retains a high proportion of tokens (0.6-0.8) in shallower layers, applies progressively greater sparsity (i.e., reduces the keep ratio) in intermediate layers, and ultimately reduces the ratio to a minimal level in the deepest layers, where nearly all tokens bypass the FFN. Furthermore, the AdaS-FFN demonstrates a pronounced tendency to adaptively adjust sparsity in the middle layers (e.g., layers 11-28), as indicated by the larger variance in this segment.

To further elucidate the fine-grained operational principles of our modules, we visualized a specific case study. As illustrated in Figure~\ref{fig:example}, we posed a temporal-grounding query to a minute-long video. For this analysis, we computed two metrics: (1) for the AdaS-Attn module, the average selection frequency of each Cube by text tokens (denoted 'Attn Slc.'), and (2) for the AdaS-FFN module, the average token keep ratio within each Cube (denoted 'FFN Slc.'). We observed that the statistics from both modules exhibit information retention patterns that strongly correlate with the temporal localization required by the query. This result further validates our modules' capability for key information extraction and their capacity for fine-grained visual-text modal interaction.
\section{Conclusion}
In this work, we introduced AdaSpark, an adaptive sparsity framework to address the prohibitive computational cost of long-form video understanding in Video-LLMs. We demonstrated that existing methods often compromise perceptual fidelity or long-range temporal modeling. Our approach, which partitions video into spatio-temporal cubes, successfully mitigates these issues through two co-designed components: Adaptive Cube-Selective Attention  and Adaptive Token-Selective FFN. By leveraging an entropy-based selection mechanism, AdaSpark adaptively allocates computation based on content complexity. Our experiments confirm that this strategy achieves significant computational reductions by up to 57\% FLOPs while preserving fine-grained details and long-range dependencies, offering comparable performance to dense models. This validates adaptive sparsity as a viable and effective path for scaling Video-LLMs to handle real-world, long-duration video inputs.

\section*{Acknowledgments}

This research is supported by the Strategic Priority Research Program of Chinese Academy of Sciences under Grant XDB1350103, and the National Natural Science Foundation of China (62437001, 62436001, 62531026), and the Natural Science Foundation of Jiangsu Province under Grant BK20243051.
{
    \small
    \bibliographystyle{ieeenat_fullname}
    \bibliography{main}
}

\clearpage
\maketitlesupplementary

\section{Effect of Selection Strategy}
Adhering to the ablation experimental setup described in the main text, we investigate the influence of various token selection strategies under an identical compression ratio. As summarized in Table \ref{tab:selection_abla}, we initially evaluate the most rudimentary approach: uniform sampling. This method exhibits the most significant performance degradation under equivalent compression levels. Subsequently, we report the performance of the static Top-K strategy, which serves as our primary comparative baseline. Due to its inability to dynamically select visual cubes across different layers, the Top-K approach lags behind our method, resulting in performance deficits of 1.9 on Charades STA, 3.1 on VideoMME, and 2.4 on MLVU. We further explore a constrained variant that applies Top-K selection exclusively to fixed I-frames (keyframes identified by traditional video compression algorithms); the results indicate that the performance deviation from the standard Top-K approach is negligible. In contrast, our AdaSpark employs a dynamic Top-P mechanism, which facilitates a more flexible selection strategy and yields superior performance.
\begin{table}[h]
\caption{\textbf{Ablation on more selection strategy.}}
\vspace{-3mm}
\label{tab:selection_abla}
\centering
\resizebox{1.0\linewidth}{!}{
\begin{tabular}{l cccc}
\toprule
 {Selection Strategy}  & Charades STA & VideoMME & MLVU \\

\midrule
 Uniform Sampling & 31.5 & 50.2 & 53.8 \\
 Top-K & 34.2 & 52.6 & 55.1 \\
 I-Frame & 34.0 & 52.9 & 55.7 \\
\rowcolor{lightblue}
 Top-P & \textbf{36.1} & \textbf{55.7} & \textbf{57.5}  \\

\midrule

\end{tabular}
}
\vspace{-5mm}
\end{table}

\section{Implementation Details}
\subsection{Training Configuration}~\label{sec:sup_detail}
Table~\ref{tab:hyperparams} provides a comprehensive summary of the hyperparameters employed in the training of AdaSpark. Throughout this process, the visual encoder is maintained in a frozen state, and we implement a cube-based sparse strategy regulated by a top-$p$ threshold.

\begin{table}[!htbp]
\centering
\caption{\textbf{Hyperparameter configuration for AdaSpark training.} The model undergoes a single-stage post-training protocol.}
\label{tab:hyperparams}
\resizebox{0.9\linewidth}{!}{
\begin{tabular}{lc}
\toprule
\textbf{Hyperparameter} & \textbf{Value} \\
\midrule
\multicolumn{2}{l}{\textit{Model Configuration}} \\
Backbone & Qwen2.5-VL-3B / 7B \\
Visual Encoder Status & Frozen \\
Max Visual Sequence Length & 48k \\
Max Context Length & 64k \\
Cube Size ($H \times W \times T$) & $8 \times 8 \times 4$ \\
Top-$p$ Threshold & 0.7 \\
\midrule
\multicolumn{2}{l}{\textit{Training Optimization}} \\
Data Scale & $\sim$255K (178K + 77K) \\
Global Batch Size & 256 \\
Learning Rate (lr) & $2 \times 10^{-6}$ \\
Sequence Parallel & 4 \\
LR Schedule & Cosine Decay \\
Optimizer & AdamW \\
Weight Decay & 0 \\
DeepSpeed Stage & Zero2 \\
\midrule
\multicolumn{2}{l}{\textit{Data \& Hardware}} \\
Input FPS & 4 \\
Compute Resources & 32 $\times$ NVIDIA H100 \\
Training Time & $\approx$ 4 Days \\
\bottomrule
\end{tabular}
}
\end{table}

Our training methodology incorporates a mixed dataset, centrally featuring the llava-video-178k~\cite{zhang2024video} dataset, which serves as a foundational corpus for basic video understanding. In addition, we augment this data with 77k timestamp-grounded samples from DideMo~\cite{anne2017localizing} and ActivityNet Captions~\cite{krishna2017dense} to enhance the model's capacity for identifying key temporal information.

\subsection{Pseudocode}
To provide a comprehensive understanding of our method, we present the pseudocode detailing the AdaS-Attn and AdaS-FFN mechanisms in Algorithm \ref{alg:adaspark_concise}.
\begin{algorithm}[t]
\small
\caption{AdaSpark Training Algorithm}
\label{alg:adaspark_concise}
\SetAlgoLined
\DontPrintSemicolon
\KwIn{Video Tokens $X_{vid}$, Text $X_{txt}$, Cube $(t,h,w)$, $p$}

\For{each transformer layer}{
    $Q, K, V \leftarrow \text{Linear}(X)$; \quad Apply $\text{RoPE}(Q, K)$\;
    \textbf{AdaS-Attn}: Reshape $K, V$ into Cubes $(B, N, t \cdot h \cdot w, D)$\;
    For each query, select Top-$p$ relevant cubes based on proxy scores\;
    Calculate Attention on \{Selected Cubes $\cup$ Local Cube\};
    
    \textbf{AdaS-FFN}: Reshape video tokens into Cubes\;
    Calculate token importance via $L_2$-norm; Select Top-$p$ salient tokens\;
    Apply FFN to selected vision tokens and full text tokens; Add Mean(FFN($Active$)) to others\;
    
    Restore original sequence shape and add residual connections\;
}
\Return Updated $X_{seq}$
\end{algorithm}

We provide the implementation of critical components for reference; please refer to \texttt{AdaS-Attn.py} and \texttt{AdaS-FFN.py} in zip file for specific details.

To facilitate a more detailed understanding of the compression achieved by our algorithm on standard causal attention and FFN layers, we provide a comprehensive discussion on the theoretical FLOPs calculation in the following section.

\subsection{Detail of FLOPs Calculation}
\paragraph{FLOPs Analysis for AdaS-Attn.} We compare the theoretical FLOPs of standard dense causal attention with our proposed AdaS-Attn. Let $S$, $D_{model}$, and $C$ denote the sequence length, model dimension, and the number of tokens per spatiotemporal cube, respectively.
For standard \textbf{dense causal attention}, computational costs arise from the Query-Key ($QK^T$) and Attention-Value ($AV$) multiplications. Since each query attends to all preceding keys/values (averaging $S/2$ tokens due to the causal mask), the total complexity exhibits a quadratic dependence on sequence length:
\begin{equation}
\begin{aligned}
FLOPs_{Dense} &\approx 2 \times S \times \frac{S}{2} \times D_{model}\\
& + 2 \times S \times \frac{S}{2} \times D_{model} \\
&\approx 2 S^2 D_{model}
\end{aligned}
\end{equation}
In contrast, \textbf{AdaS-Attn} computes attention sparsely. Let $\bar{N}$ be the average number of top-$p$ cubes selected per query. The attention mechanism is restricted to the tokens within these selected cubes (totaling $\bar{N} \times C$ tokens). The complexity for the sparse $QK^T$ and $AV$ phases is calculated as:
\begin{equation}
\begin{aligned}
FLOPs_{AdaS} &= FLOPs_{QK^T} + FLOPs_{AV} \\
&\approx (2 S \cdot \bar{N} C \cdot D_{model})\\
&+ (2 S \cdot \bar{N} C \cdot D_{model}) \\
&\approx \mathbf{4} S \bar{N} C D_{model}
\end{aligned}
\end{equation}
Since the number of selected tokens is significantly smaller than the full sequence ($\bar{N} C \ll S$), AdaS-Attn achieves linear complexity $O(S)$, drastically reducing computational overhead compared to the $O(S^2)$ standard attention.

\paragraph{FLOPs Analysis for AdaS-FFN.} We further analyze the efficiency gains in the Feed-Forward Network (FFN). Let $D_{ff}$ denote the intermediate dimension of the FFN (typically $4D_{model}$ or similar).In a Standard FFN, every token in the sequence $S$ undergoes projection up to $D_{ff}$ and down to $D_{model}$. The total FLOPs are dominated by these dense matrix multiplications:
\begin{equation}
\begin{aligned}
FLOPs_{Std-FFN} &\approx 2 \times S \times D_{model} \times D_{ff}\\
&+ 2 \times S \times D_{ff} \times D_{model}\\
&= 4 S D_{model} D_{ff}
\end{aligned}
\end{equation}
For our AdaS-FFN, computation is content-aware. The additional overhead for calculating $L_2$-norms and the importance distribution $S_i$ is $O(S \cdot D_{model})$, which is negligible compared to the matrix transformations ($D_{model} \times D_{ff}$).The heavy FFN computation is only applied to the set of activated tokens $M_i$. Let $S_{act} = \sum |M_i|$ be the total number of activated tokens across all cubes, and $\bar{r} = S_{act}/S$ be the average activation ratio. The inactive tokens $R_i$ bypass the FFN and utilize the Mean Compensation strategy, which involves only lightweight vector addition operations ($O(S \cdot D_{model})$).Thus, the FLOPs for AdaS-FFN are proportional only to the activated tokens:
\begin{equation}
\begin{aligned}
FLOPs_{AdaS-FFN} &\approx 4 \times S_{act} \times D_{model} \times D_{ff} \\
&= \mathbf{4} \times (\bar{r} \cdot S) \times D_{model} \times D_{ff}
\end{aligned}
\end{equation}
Given that $\bar{r}$ is controlled by the top-$p$ threshold (typically $\bar{r} \ll 1$), AdaS-FFN significantly reduces the FLOPs by a factor of $\bar{r}$ compared to the standard FFN, while preserving semantic integrity through mean compensation.

\section{Evaluation Settings}
We evaluate AdaSpark on a series of comprehensive video-language benchmarks: 1) Extra Long Video Understanding, using Video Needle in a Haystack~\cite{zhaoneedle,zhang2024long}; 2) Long Video Understanding, which includes MLVU~\cite{zhou2024mlvu}, VideoMME~\cite{fu2024video}, LongVideoBench~\cite{wu2024longvideobench}, and LVBench~\cite{wang2025lvbench}; 3) Short Video Understanding, using MVBench~\cite{li2023mvbench}; 4) Spatial Reasoning, with VSIBench~\cite{yang2025thinking}; and 5) Video Grounding, utilizing CharadesSTA~\cite{gao2017tall}. Table~\ref{tab:eval_settings} details the frame sampling configurations employed during inference via the \texttt{lmms-eval} framework. For all tasks, we strictly adhere to the default prompts and scoring protocols provided by the evaluation framework.

\begin{table}[!htbp]
\centering
\caption{\textbf{Evaluation settings summary for each benchmark.} For all benchmarks, we set the temperature, top $p$, and number of beams to 0, 0, and 1, respectively. \textbf{FPS} denotes the sampling frames per second, and \textbf{\# F} represents the maximum number of sampling frames allowed.}
\label{tab:eval_settings}
\resizebox{0.85\linewidth}{!}{
\begin{tabular}{lcc}
\toprule
\textbf{Benchmark} & \textbf{FPS} & \textbf{\# F (Max Frames)} \\
\midrule
VideoNIAH & 1 & 4096 \\
\midrule
MLVU (Dev) & 1 & 512 \\
VideoMME & 1 & 256 \\
LongVideo & 1 & 1024 \\
LVBench & 1 & 1024 \\
\midrule
MVBench & 2 & 256 \\
VsiBench & 2 & 256 \\
CharadesSTA & 4 & 256 \\
\bottomrule
\end{tabular}
}
\end{table}

\end{document}